\documentclass{article}
\pdfoutput=1
\usepackage{fullpage}
\usepackage[T1]{fontenc}
\usepackage[light,math]{iwona}

\usepackage{amsthm} 
\usepackage{amssymb}  
\usepackage{amsmath} 
\usepackage{mathrsfs}
\usepackage{txfonts}
\usepackage{graphicx}
\usepackage{color}
\usepackage{multirow}
\usepackage{epstopdf}
\usepackage{subfig}

\newcommand{\E}{{\mathbb{E}}}

\newcommand{\Dxitr}{{D_x}^{i,train}}
\newcommand{\Dxonetr}{{D_x}^{1,train}}

\newcommand{\Dxite}{{{D_x}^{i,test}}}

\newcommand{\fhat}{{\hat{f}}}
\newcommand{\fhatis}{{\hat{f}_{is}}}
\newcommand{\ghat}{{\hat{g}}}
\newcommand{\ghati}{{\hat{g}_i}}
\newcommand{\fhatmc}{{\hat{f}_{mc}}}
\newcommand{\fhatsmc}{{\hat{f}_{smc}}}

\definecolor{Red}{rgb}{0.9,0.0,0.0}

\begin{document}
\title{Using Supervised Learning to Improve Monte Carlo Integral Estimation}
\author{Brendan Tracey 
 \\ \normalsize\itshape{Stanford University, Stanford, CA 94305}
\\ David Wolpert 
\\ \normalsize\itshape{NASA Ames Research Center, Moffett Field, CA 94035} \\ 
Juan J. Alonso 
 \\ \normalsize\itshape{Stanford University, Stanford, CA 94305}}
\maketitle

\begin{abstract}
Monte Carlo (MC) techniques are often used to estimate integrals of a multivariate function using randomly generated samples of the function. In light of the increasing interest in uncertainty quantification and robust design applications in aerospace engineering, the calculation of expected values of such functions (e.g. performance measures) becomes important. However, MC techniques often suffer from high variance and slow convergence as the number of samples increases.
In this paper we present Stacked Monte Carlo (StackMC), a new method for post-processing an existing set of MC samples to improve the associated integral estimate. StackMC is based on the supervised learning techniques of fitting functions and cross validation. It should reduce the variance of any type of Monte Carlo integral estimate (simple sampling, importance sampling, quasi-Monte Carlo, MCMC, etc.) without adding bias. We report on an extensive set of experiments confirming that the StackMC estimate of an integral is more accurate than both the associated unprocessed Monte Carlo estimate and an estimate based on a functional fit to the MC samples. These experiments run over a wide variety of integration spaces, numbers of sample points, dimensions, and fitting functions. In particular, we apply StackMC in estimating the expected value of the fuel burn metric of future commercial aircraft and in estimating sonic boom loudness measures. We compare the efficiency of StackMC with that of more standard methods and show that for negligible additional computational cost significant increases in accuracy are gained.

\end{abstract}

\section*{Nomenclature}
 \begin{tabbing}
  $b$ ~~~~~~~~~~~~~~~\= Bias of M \\
  $D_x$ \> Set of samples of $f(x)$ \\
  $D_x(i)$ \> $i^{th}$ Sample of $f(x)$ \\
  $\Dxite$ ~\> Subset of Samples in the $i^{th}$ Testing Set \\
  $\Dxitr$ \> Subset of Samples in the $i^{th}$ Training Set \\
  $\E[\cdot]$ \> Expected Value \\
  $f(x)$ \> Objective Function \\
  $\fhat$ \> True Expected Value of $f(x)$ \\
  $\hat{f}_{M}$ \> Estimate of $\fhat$ from M \\
  $f(D_x(i))$ \> Function Value at the $i^{th}$ sample \\
  $g(x)$ \> Fitting Algorithm \\
  $g_i(x)$ \> $i^{th}$ Fit to $f(x)$ \\
  $g_i \big(D_x(i) \big)$ \> Prediction of Fit $g_i$ at $D_x(i)$ \\
	$\ghat$ \> Expected Value of $g(x)$ \\
  $k$ \> Number of Folds \\
  $L_i$ \> Likelihood of the Expected Value of the $i^{th}$ Fold \\
  $M$ \> An Estimator of $\fhat$ \\
  $m_i$ \> Number of Samples in the $i^{th}$ Testing Set \\
  $N$ \> Number of Samples \\
  $N_i$ \> Number of Samples in the $i^{th}$ Training Set \\
  $p(x)$ \> Probability Distribution of $x$ \\
  $q(x)$ \> Alternate Sample Distribution \\
  $r(x)$ \> Fitting Algorithm for $p(x)$ \\
  $v$ \> Variance of M \\
  $x$ \> Input Parameters \\
  $\alpha$ \> Free Parameter of StackMC \\
  $\beta$ \> Free Parameter in the Fitting Algorithm \\
  $\rho$ \> Correlation \\
  $\sigma$ \> Standard Deviation \\ 
 \end{tabbing}

\section{Introduction}
A system optimized only for best performance at \emph{nominal} operating conditions can see severe performance degradation at even slightly \emph{off-design} conditions. Aerospace systems rarely operate at precisely the design condition and a robust design approach dictates trading some performance at the nominal condition for improved performance over a wide range of input conditions. However, certain input conditions will occur rarely or never, and adding robustness for these conditions will degrade average system performance. Consequently, we are often interested in optimizing the \emph{expected} performance of a system rather than the performance at any specific operating point. 

Formally, we are interested in an integral of the form
\begin{equation}
\label{eq:EV}
\E[f]=\fhat = \int f(x)p(x)\,dx \quad \text{,}
\end{equation} 
where $f(x)$ is the (potentially multivariate) objective function to be optimized, and $p(x)$ is the probability density function (or probability distribution, in which case \eqref{eq:EV} is actually a summation) from which $x$ is generated.

Evaluating \eqref{eq:EV} is non-trivial, especially when the objective function is high dimensional and/or expensive to evaluate. When standard quadrature techniques (such as Simpson's rule) become intractable, Monte Carlo (MC) methods are powerful techniques which have become the standard for integral estimation. However, MC techniques also have their drawbacks; the required computational expense may be prohibitive in most situations. The question is if a technique exists to significantly lower the cost of estimating \eqref{eq:EV}. This paper describes such a technique which combines MC with machine learning and statistical techniques and can lead to significant computational savings over traditional methods. 

We begin this paper with a brief review of integral estimation techniques including Monte Carlo, fitting algorithms, and stacking. We then introduce a new technique we call Stacked Monte Carlo (StackMC), which uses stacking to reduce the estimation error of any Monte Carlo method. Finally, we apply StackMC to a number of analytic example problems and to two problems from the aerospace literature. We show that StackMC can significantly reduce estimation error and never has higher estimation error than the best of Monte Carlo and the chosen fitting algorithm.
\section{Integral Estimation Techniques}
\subsection{Estimation Error}
When using any method $M$ returning $\hat{f}_M$ as an estimate of \eqref{eq:EV}, there are two sources of estimation error: bias ($b$) and variance ($v$). Bias is the expected difference between the method and the truth: $b=\E[\hat{f}_M-\fhat]$, where the expectation is over all data sets. A method whose average over many runs gives the correct answer has zero bias, whereas a method that estimates high or low on average has non-zero bias. As an example, the Euler equations have significant bias in their estimate of viscous drag. Variance is a measure of how much $\hat{f}_M$ varies between different runs of $M$: $v=\E[(\hat{f}_M-\E[\hat{f}_M])^2]$. If $M$ is deterministic, it has zero variance because every run returns the same answer, whereas if $M$ is stochastic multiple runs have different outputs leading to a non-zero variance.  The total expected squared error is the combination of these two factors
\begin{equation}
\nonumber
\E \big{[}|\hat{f}_M - \hat{f}|^2 \big{]} = b^2+v \quad .
\end{equation}

Any method seeking to reduce estimation error must keep both of sources of error low.

\subsection{Monte Carlo Techniques}
 
In ``Simple Sampling'' Monte Carlo (MC), a set of $N$ samples, $D_x$, is generated randomly according to $p(x)$. The estimate of the expected value of the function based on $D_x$ is the average of the function values of the samples
\begin{equation}
\fhat \approx \fhatmc = \frac{1}{N} \sum_{i=1}^{N}f\big{(}D_x(i)\big{)} \quad \text{,}
\nonumber
\end{equation}
where $D_x(i)$ refers to the $i^{th}$ generated sample. Simple Monte Carlo has two extremely useful properties. First, MC is guaranteed to be unbiased and thus, on average, will return the correct answer. Second, MC is guaranteed to converge to the correct answer at a rate of $O(n^{1/2})$ \emph{independent of the number of dimensions}. That is, for \emph{any} problem, increasing the number of samples by a factor of one hundred decreases the expected error by a factor of ten.  As Simple Monte Carlo has zero bias, all of the error comes from the variance of Monte Carlo runs. This variance is due a different kind of variance concerning $f(x)$ itself; fluctuations in the Monte Carlo estimate are directly related to fluctuations in $f(x)$. As a result, the smaller the variance of $f(x)$, the smaller the expected error of Monte Carlo.

Many different sample generation techniques exist to help reduce the variance of Monte Carlo. Importance sampling \cite{HaughII} generates samples from an alternate distribution $q(x)$ (instead of the true distribution $p(x)$) and estimates integral \eqref{eq:EV} as a weighted average of sample values 
\begin{equation}
\nonumber
\fhat=\int f(x)p(x)\,dx= \int \frac{f(x)p(x)}{q(x)}q(x)\,dx \quad \text{,} 
\end{equation}
\begin{equation}
\label{eq:IS}
\fhatis=\frac{1}{N} \sum_{i=1}^{N}\frac{f\big{(}D_x(i)\big{)}p\big{(}D_x(i)\big{)}}{q\big{(}D_x(i)\big{)}} \quad \text{.}
\end{equation}
Importance sampling is often used when the tails of a distribution have a measurable effect on $\fhat$, but occur very infrequently. Using Simple MC, several million samples are needed to accurately capture a once-in-a-million event, but by using importance sampling unlikely outcomes can be made to occur more frequently reducing the total number of samples needed.

Quasi-Monte Carlo (QMC) techniques reduce variance by choosing sample locations more carefully. Sample points generated from simple Monte Carlo will inevitably cluster in some locations and leave other locations void of samples. QMC methods are usually deterministic and reduce variance by spreading points evenly throughout the sample space. Two such methods are the scrambled Halton sequence \cite{Halton} and the Sobol sequence \cite{Sobol}. While often effective, due to deterministic sample generation they are not guaranteed to be unbiased. It is also difficult to generate points from an arbitrary $p(x)$; most QMC algorithms generate sample points from a uniform distribution over the unit hypercube.


\subsection{Supervised Learning}   
A second class of techniques for estimating integrals from data seeks to use the data samples more efficiently through the use of a fitting algorithm and supervised learning techniques. The fitting algorithm uses data samples to make a ``fit'' to the data, and the integral estimate is the integral of the fit. Examples of fitting algorithms include splines, high-order polynomials and Fourier expansions; even Simpson's rule computes the quadrature by fitting a piecewise quadratic polynomial to the data samples. Fits incorporate the spatial distribution of sample points and thus often give more accurate estimates of $\fhat$ than MC. However, using a fitting algorithm can induce bias and may or may not exhibit convergence to the correct answer as the number of points increases. Additionally, when the number of data points is ``too small'' many fitting algorithms exhibit higher variance than MC, and, as a result, using a fitting algorithm can be worse than not using one, especially with low numbers of sample points and in high-dimensional spaces.  It is usually impossible to know \emph{a priori} how many points is ``too small'' making it difficult to know when to use a fitting algorithm. 

Additionally, it is difficult to know if a fit to the data samples is an accurate representation of $f(x)$ at values of $x$ not in the data set. Many fitting algorithms exhibit a property known as ``overfitting'' where a fit to the data is very accurate at $x$ locations that are among the data samples, but is very inaccurate at $x$ locations not among the data samples. As a result, one cannot use the data samples themselves to both make a fit and to evaluate its accuracy. The standard technique for addressing this issue is known as cross-validation, where a certain percentage of the data is used to make the fit and the rest of the data is used to evaluate its performance. Usually this process is repeated several times, and the best performing fit is used as the output of the fitting algorithm (a.k.a. \emph{winner-takes-all}).

Stacking is a technique first introduced by Wolpert \cite{Wolpert:Stacking} which improves upon cross-validation by combining the results of different fits. Wolpert describes stacking in his paper as `` a strategy ... which is more sophisticated than winner-takes-all. Loosely speaking, this strategy is to combine the [fits] rather than choose one amongst them. ... [Stacking] can be viewed as a more flexible version of nonparametric statistics techniques like cross-validation, ... therefore it can be argued that for almost \emph{any} generalization or classification problem ... one should use [stacking] rather than any single generalizer by itself.'' 
Interested readers can see stacking applied to improving regression \cite{BreimanSR}, probability density estimation \cite{Smyth}, confidence interval estimation \cite{Kim}, and optimization \cite{Dev}. Stacking recently gained fame as being a major part of the first and second place algorithms in the Netflix competition \cite{Sill}.

In addition to combining the predictions of multiple fitting
algorithms, stacking can also be used to improve the prediction of a single fitting algorithm. In this variant of stacking, one again partitions the full set of data $D_x$ into a subset $S_1$ used to make fits and the remainder of the data, and a subset $S_2 = D_x \setminus S_1$. However now
one does not use $S_2$ to determine how to combine the predictions of multiple fitting algorithms that were all run on $S_1$. Instead one uses $S_2$ to correct the prediction of a single algorithm that was run on $S_1$. This variant
of stacking is closest to the algorithm we use below. For a comparison of stacking to other generalizers, see Clarke\cite{Clarke}.

\section{Stacked Monte Carlo}

The main idea of Stacked Monte Carlo (StackMC) is to incorporate the variance reduction potential of a fitting algorithm while avoiding introducing bias and additional variance from poor fits. It makes no assumptions about the function $f(x)$ nor the sample generation method, and therefore StackMC can be used to augment nearly any Monte Carlo application.

Let us assume that we have a function $g(x)$ that is a reasonable (though not necessarily perfect) fit to $f(x)$. Equation \eqref{eq:EV} can be re-written as 

\begin{eqnarray}
\label{eq:smc1_nofold}
\hat{f} &= & \int{\alpha g(x)p(x)}\,dx + \int{\big{(}f(x)-\alpha g(x)\big{)}p(x)}\,dx \nonumber \\
        &= & \hspace*{0.37in} \alpha \hat{g}                          \hspace*{0.37in}+\int{\big{(}f(x)-\alpha g(x)\big{)}p(x)}\,dx \quad \text{,}
\end{eqnarray}
where $\alpha$ is a constant and $\ghat= \int g(x) p(x)\,dx$. Instead of using Monte Carlo samples to estimate \eqref{eq:EV} directly, we can use the Monte Carlo samples to estimate the integral term in \eqref{eq:smc1_nofold}, i.e.

\begin{equation}
\label{eq:smc1_nofold_sum}
\hat{f} \approx \alpha \ghat + \frac{1}{N} \sum_{i=1}^N{f\big{(}D_x(i)\big{)}-\alpha g\big{(}D_x(i)\big{)}} \quad .
\end{equation}

\begin{figure}[h!]
\centering
\subfloat[Example $f(x)$ and $g(x)$]{\includegraphics[width=0.4\textwidth]{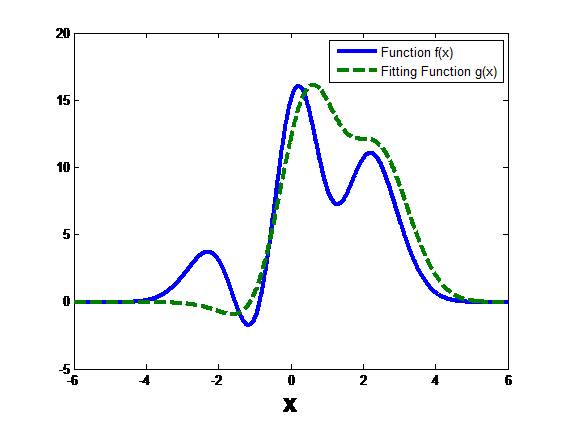}}
\subfloat[$f(x)$ and $f(x)-\alpha g(x)$]{\includegraphics[width=0.4\textwidth]{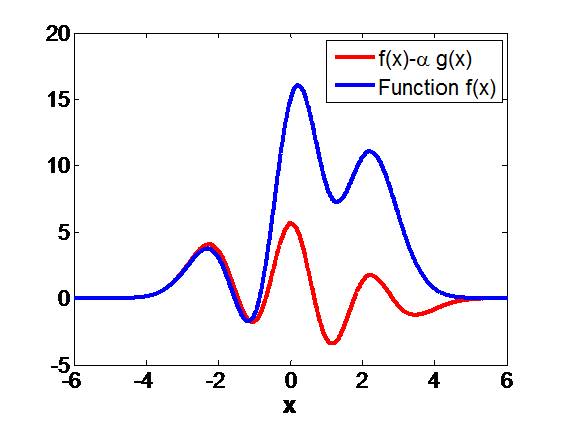}}
\caption{Comparison of $f(x)$ and $f(x)-\alpha g(x)$ for a value of $\alpha=0.85$. $f-\alpha g$ has lower variance than $f$ alone, and so a Monte Carlo estimate of $f-\alpha g$ will have less error than an estimate of $f$.}
\label{fig:VarRed}
\end{figure}

Since $g(x)$ is assumed to be a reasonable fit, for a properly chosen $\alpha$, $f(x)-\alpha g(x)$ has lower variance than $f(x)$ alone (see Fig.~\ref{fig:VarRed}), and so a MC estimate of \eqref{eq:smc1_nofold} has lower expected error than an estimate of \eqref{eq:EV}. In fact, it can be shown \cite{HaughI} that the optimal value of $\alpha$ to minimize expected error is 
\begin{equation}
\label{eq:smc1alpha*}
\alpha=\rho \frac{\sigma_f}{\sigma_g} \quad \text{,}
\end{equation}
where $\sigma_f$ and $\sigma_g$ are the standard deviations of $f(x)$ and $g(x)$ respectively, and $\rho$ is the correlation between $f$ and $g$. Intuitively, if $g$ is a good fit to $f$, $\rho$ (and correspondingly $\alpha$) will be high and $\ghat$ will be trusted as a good estimate for $\fhat$. If $g$ is a poor fit to $f$, $\rho$ and $\alpha$ will both be low and $\ghat$ will be downplayed.  Looking at it from a different perspective, \eqref{eq:smc1_nofold} takes the expected value estimated from $g(x)$, and corrects it with a term estimating the bias of $g(x)$. Either way, by using \eqref{eq:smc1_nofold} the error should be lower than using either Monte Carlo or the fitting function alone. Since the expected value of the fit is both added and subtracted, \eqref{eq:smc1_nofold_sum} remains an unbiased estimate of \eqref{eq:EV}. Thus, \eqref{eq:smc1_nofold_sum} incorporates a fitter while remaining unbiased, and $\alpha$ allows us to emphasize good fits while de-emphasizing poor ones. 

The obvious question is how to obtain $g(x)$ and find $\alpha$ from data samples. The first step is to pick a functional form for a fitting algorithm $g(x)$, such as a polynomial with unknown coefficients. $g(x)$ can be nearly anything, the only restrictions are that it can make predictions at new $x$ values. For now we also require that $g$ can be analytically integrated over the volume, i.e. we can calculate

\begin{equation}
\label{eq:ghat}
\ghat= \int g(x) p(x)\,dx
\end{equation}
analytically (see further discussion later in the paper). By comparing the output of a fit $g(x)$ to the true $f(x)$ values, an estimate for the ``goodness'' of the fit (and by extention $\alpha$) could be obtained, and \eqref{eq:smc1_nofold_sum} could be applied. However, doing this directly would cause over-fitting and would lead to an inaccurate estimate of $\alpha$.

Over-fitting can be mitigated by using a technique known as $k$-fold cross-validation \cite{DudaHartStork}. The $N$ data samples are randomly partitioned into $k$ testing sets, $\Dxite, i=1,...,k$, which are mutually exclusive and contain the same number of samples (differing slightly if $N/k$ is not an integer). Training sets, $\Dxitr, i=1,...,k$, contain all of the data samples not in $\Dxite$. We call $N_i$ the number of samples in $\Dxite$ and $m_i$ the number of samples in $\Dxitr$ (such that $N_i + m_i = N$). One fit per fold, $g_i(x)$, is created using only the $N_i$ samples in $\Dxitr$ (for a total of $k$ fitters). The samples in a training set are called ``held in'' points because they are used to generate the fit, whereas points in the corresponding testing set are ``held out'' points because the fit is generated without using these samples. Each data sample is in a held out data set once and in the held in data set $k-1$ times. Using $g_i(x)$ a prediction of the function values for the points in $\Dxite$ is made ( $g_i(\Dxitr)$ ). Standard cross-validation evaluates the accuracy of each of the fits, and chooses the best fit to use as a single $g(x)$. 

Instead, we adopt the approach of stacking and use \eqref{eq:smc1_nofold} to get an estimate of $\fhat$ from each of the fitters, and we can use the held out data points as an estimate of the integral term.

\begin{align}
\label{eq:fitterint}
\fhatsmc(i) =& \alpha \ghat_i+ \int{\big(f(x)-\alpha g_i(x) \big)p(x)}\,dx \\
\label{eq:fittersum}
\fhatsmc(i)=&\alpha \hat{g_i}  + \frac{1}{m_i} \sum_{j=1}^{m_i}{f \big{(}\Dxite(j) \big{)}-\alpha g_i \big{(}\Dxite(j) \big{)}} \quad \text{.}
\end{align}
The mean of these estimates is taken as the final prediction
\begin{equation}
\label{eq:fittercombine}
\fhat \approx \fhatsmc = \frac{1}{k} \sum_{i=1}^k \fhatsmc(i) \quad \text{.}
\end{equation}

We can also use the predictions at the held out points to estimate $\alpha$. $\sigma_f$ is estimated from the variance of the sample function values themselves, $\sigma_g$ from the variance of the predictions, and $\rho$ from the correlation between the predictions and the truth.

\begin{align}
\nonumber
&\mu_f=\frac{1}{N}\sum_{j=1}^N f\big{(}D_x(j)\big{)} \\
\nonumber
&\mu_g=\frac{1}{N}\sum_{i=1}^k \sum_{j=1}^{N_i} g_i\big{(}\Dxitr(j)\big{)} \\
\nonumber
&\sigma_f=\sqrt{\frac{1}{N-1}\sum_{j=1}^{N} \bigg{(}f\big{(}D_x(j)\big{)}-\mu_f \bigg{)}^2} \\
\nonumber
&\sigma_g=\sqrt{\frac{1}{N-1}\sum_{i=1}^k \sum_{j=1}^{N_i} \bigg{(}g_i\big{(}\Dxitr(j)\big{)}-\mu_g\bigg{)}^2} \\
\nonumber
&cov(f,g)=\frac{1}{N-1}\sum_{i=1}^k \sum_{j=1}^{N_i} \bigg{(}f\big{(}\Dxitr(j)\big{)}-\mu_f\bigg{)}\bigg{(}g\big{(}\Dxitr(j)\big{)}-\mu_g\bigg{)} \\
\nonumber
&\rho=\frac{cov(f,g)}{\sigma_f \sigma_g} \quad \text{.}
\end{align}

Finally, we plug into \eqref{eq:fittercombine}

\begin{equation}
\label{eq:smc1}
\fhatsmc=\frac{1}{k}\sum_{i=1}^k \fhatsmc(i) =\frac{1}{k}\sum_{i=1}^k{\bigg{[}\alpha \hat{g_i}  + \frac{1}{m_i} \sum_{j=1}^{m_i}{f \big{(}\Dxite(j) \big{)}-\alpha g_i \big{(}\Dxite(j) \big{)}}\bigg{]}} \quad \text{.}
\end{equation}

One final correction is needed to complete StackMC. It is possible that some or all of the $\ghati$ differ greatly from $\fhat$ but $\alpha$ is calculated high because of good predictions at held out data points. If left alone, this would cause StackMC to return a low-quality prediction on some data sets.

However, the error in the mean (EIM) of the Monte Carlo samples can be used as a second metric to evaluate the goodness of the fit. EIM is defined as
\begin{equation}
\nonumber
\bar{\sigma}=\frac{\sigma}{\sqrt{N}} \quad ,
\end{equation}
and represents the uncertainty in the mean of a set of Monte Carlo samples. Specifically, it gives us a likelihood bound on $\fhat$ based on $\fhatmc$ and $\sigma_f$. We can find a ``likelihood'' for each fold by calculating
\begin{equation}
\tilde{L}_{i}=\frac{| \ghati - \fhatmc |}{\bar{\sigma}} \quad .
\end{equation}
The higher $L_i$, the less likely it is that $\ghati=\fhat$, and the more likely it is that the fitter is bad and should be ignored. A heuristic is set that if $\max(L_i)>C$, $\fhatsmc=\fhatmc$, i.e. all of the fits are ignored entirely and the Monte Carlo average is used.  If $C$ is set too low, then too many reasonable fits are ignored, and if $C$ is set too high, then too many unreasonable fits are kept. A value of $C=5$ was chosen based on experimentation; it was clear that setting $C$ as low as $3$ or as high as $7$ were inferior.

Finally, a note about the bias of StackMC. Because the expected value of the fit is being both added and subtracted, it is true that \eqref{eq:smc1_nofold} (and by extension, \eqref{eq:fitterint} and \eqref{eq:fittercombine}) are also unbiased for a fixed $\alpha$. However, using the held-out data to both set $\alpha$ and estimate the integral can introduce bias. In practice, we have found that this is only a problem for very small numbers of sample points where the output fit of the fitting algorithm changes significantly depending on the held-in samples.

\subsection{Generalized Stacked Monte Carlo}
The discussion above was for the case where the samples are generated according to a known $p(x)$, this is only a special case of a class of estimation scenarios. While the overall methodology remains approximately the same for these other scenarios, some specifics (such as the calculation of $\alpha$) change with the choice for $g(x)$ and the sample generation method.

\subsubsection{Importance sampling}
If importance sampling methods are used, samples are generated from $q(x)$ instead of $p(x)$. As a result, \eqref{eq:EV} is expanded as
\begin{equation}
\nonumber
\hat{f} = \int{\alpha g(x) q(x)}\,dx + \int{\bigg{(}\frac{f(x)p(x)}{q(x)}-\alpha g(x)\bigg{)}q(x)}\,dx \nonumber \quad \text{,}
\end{equation}
and so $g(x)$ should be a fitting algorithm for $\frac{f(x)p(x)}{q(x)}$ instead of just $f(x)$. As a result a few modifications are needed including the fact that, 

\begin{equation}
\nonumber
\ghati = \int g(x) q(x) \,dx \quad \text{,}
\end{equation}
and that in the calculation of $\alpha$ and \eqref{eq:smc1}, $\frac{fp}{q}$ is used instead of just $f$.

\subsubsection{Unable to integrate $g(x)$ over $p(x)$}
If the samples are generated from $p(x)$, but the choice for $g(x)$ cannot be integrated over $p(x)$, there are two options. The first option is to use another function $r(x)$ as a fitting algorithm for $p(x)$ over which $g(x)$ is integrable. We expand \eqref{eq:EV} as

\begin{eqnarray}
\nonumber
\hat{f} &=& \int{\alpha g(x)r(x)}\,dx + \int{f(x) p(x) - \alpha g(x)r(x)}\,dx \\
\nonumber
&=& \int{\alpha g(x)r(x)}\,dx + \int{f(x) - \alpha \frac{g(x)r(x)}{p(x)} p(x)}\,dx \quad \text{.}
\end{eqnarray}
Similarly to the above case, when calculating $\alpha$ and \eqref{eq:smc1}, $g$ is replaced with $\frac{gr}{p}$.

Alternately, when $g(x)$ is significantly less computationally expensive than $f(x)$, $\ghat$ itself can be estimated by Monte Carlo techniques. When estimating $\ghat$ from $N_g$ additional samples, it should be the case that $N_g \gg N$ so that errors in the estimate of $\ghat$ do not cause significant errors in the estimate of $\fhat$.


\subsection{A Simple Example}

We attempt to find the expected value of $3x^6 + 3.6 x^5 -91.29 x^4-19.41x^3+57.15 x^2-14.43x+0.9$, where $x$ is generated according to a uniform distribution between $-3$ and $3$. Thus, the integral of concern is

\begin{equation}
\nonumber
\hat{f}=\frac{1}{2} \int_{-3}^{3} \big{(}x^6 + 1.2 x^5 -30.43 x^4-6.47x^3+19.05 x^2-4.81x+0.3\big{)}\,dx \quad \text{.}
\end{equation}

The actual value of $\fhat$ can be calculated to be 0.7069. Twenty samples are generated and divided into five folds, with each fold's testing set containing four points and each corresponding training set containing the other sixteen points. While the function of interest is a sixth order polynomial, $g(x)$ was chosen to be a third-order polynomial: $\beta_0 +\beta_1 x + \beta_2 x^2 + \beta_3 x^3$. $g_1(x)$ is found by choosing the $\beta$'s which minimize the squared error of the data in $\Dxonetr$ (using the standard pseudo-inverse). $g_2(x)$ - $g_5(x)$ are set similarly. Next, $g_i \big{(}\Dxite(j) \big{)}$ is evaluated for each test point in $\Dxite$.
\begin{table}
\centering
\begin{tabular}{|c*{8}{|r}|}
\hline
Left-out Fold & $x$ & $f(x)$ &$\beta_0$&$\beta_1$&$\beta_2$&$\beta_3$& $g_i(x)$&$\ghati$ \\
\hline
\multirow{4}{*}{1} & 0.4087 & 0.2438 &\multirow{4}{*}{2.7385}&\multirow{4}{*}{-4.3737}&\multirow{4}{*}{-3.9500}&\multirow{4}{*}{-9.4712}& -0.3549&\multirow{4}{*}{1.4281} \\
&-0.6950&7.8350&&&&& 7.0498&\\
&-0.0943&0.9259&&&&&3.1237&\\
&0.1152&-0.0166&&&&&2.1675&\\
\hline
\multirow{4}{*}{2}& 0.4117&0.2420&\multirow{4}{*}{2.4683}&\multirow{4}{*}{-3.3829}&\multirow{4}{*}{-3.0183}&\multirow{4}{*}{-11.3595}&-0.2284&\multirow{4}{*}{1.4622}\\
&0.2745&0.1108&&&&&1.0774&\\
&0.1823&-0.0163&&&&&1.6825&\\
&0.2882&0.1342&&&&&0.9708&\\
\hline
\multirow{4}{*}{3} &-0.6318&7.6689&\multirow{4}{*}{0.7900}&\multirow{4}{*}{0.3755}&\multirow{4}{*}{3.3397}&\multirow{4}{*}{-22.0257}&7.4398&\multirow{4}{*}{1.9032}\\
&-0.3923&4.7811&&&&&2.4864&\\
&-0.8345&6.4358&&&&&15.6002&\\
&0.7716&-5.2874&&&&&-7.0489&\\
\hline
\multirow{4}{*}{4} &0.5711&-0.5683&\multirow{4}{*}{1.8054}&\multirow{4}{*}{-6.7386}&\multirow{4}{*}{-0.2738}&\multirow{4}{*}{-1.3468}&-2.3831&\multirow{4}{*}{1.7141}\\
&-0.5988&7.4412&&&&&6.0312&\\
&0.9607&-16.6302&&&&&-6.1154&\\
&-0.6411&7.7172&&&&&6.3677&\\
\hline
\multirow{4}{*}{5} &0.7124&-3.2834&\multirow{4}{*}{2.3965}&\multirow{4}{*}{-3.8653}&\multirow{4}{*}{-3.4306}&\multirow{4}{*}{-10.9995}&-6.0740&\multirow{4}{*}{1.2529}\\
&0.1206&-0.0208&&&&&1.8609&\\
&-0.3960&4.8377&&&&&4.0722&\\
&0.2816&0.1230&&&&&0.7901&\\
\hline
\end{tabular}
\caption{Details of simple example calculations.}
\label{tab:simpcalc}
\end{table}

 The following parameters are calculated to find $\alpha$:
\begin{align}
\nonumber
&\mu_f= 1.1337 \\
\nonumber
&\mu_g= 1.9258\\
\nonumber
&\sigma_f= 5.6835\\
\nonumber
&\sigma_g=5.2678\\
\nonumber
&cov(f,g)=24.0999 \\
\nonumber
&\rho=\frac{cov(f,g)}{\sigma_f \sigma_g}=0.8049 \\
\nonumber
&\alpha=\rho \frac{\sigma_f}{\sigma_g}= 0.8685 \quad \text{.}
\end{align}

Finally, equation \eqref{eq:smc1} is applied to calculate $\fhatsmc$. 
\begin{table}
\centering
\begin{tabular}{|c|r|r|r|}
\hline
Fold & $\ghati$ & $ \sum_{j=1}^{m_i}{f \big{(}\Dxite(j) \big{)}-\alpha g_i \big{(}\Dxite(j) \big{)}} $ & Corrected $\ghati$\\
\hline
1&1.4281&-0.3553&0.8795\\
2&1.4622&-0.6428&0.6271\\
3&1.9032&-0.6122&1.0407\\
4&1.7141&-1.3569&0.1318\\
5&1.2529&-0.2732&1.3613\\
\hline
\multicolumn{3}{|r|}{$\fhatsmc$}&0.8081\\
\hline
\end{tabular}
\caption{Details of fold combination calculations.}
\label{tab:foldcombine}
\end{table}

\begin{figure}
\centering
\includegraphics[width=0.8\textwidth]{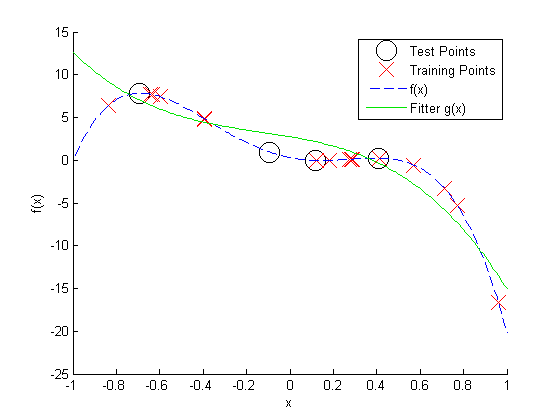}
\caption{Example of the first fold.}
\label{fig:OneFoldPicture}
\end{figure}

Using the Monte Carlo estimate alone gives $\hat{f}_{mc}=1.1337$, and using a fit to all of the data points alone gives $\hat{f}_{fit}=1.3412$. By using StackMC, we find $\fhatsmc=0.8081$, much closer to the true value than either of the individual estimates. Details of the calculations can be seen in Tables \ref{tab:simpcalc} and \ref{tab:foldcombine}, and the fit from the first fold can be seen in Fig. \ref{fig:OneFoldPicture}.

\pagebreak
\section{Results}
We test the performance of StackMC on a number of different problems. In the first set of example cases the problems have known analytic results and an exact analysis of the performance of StackMC is examined. In the second set, StackMC is applied to two aerospace problems in the literature. While an exact solution to the aerospace problems is not available, an approximate answer is obtained from a Monte Carlo estimate using 100,000 samples. For each example problem, a range of number of samples was tested using two thousand simulations at each value of $N$. In each case, 10-fold cross-validation was used. Unless otherwise noted, the plots in this section have three lines which show the mean squared error versus the number of sample points. The lines represent the average squared error from using just the prediction of Monte Carlo (green), the fit to all the samples (red), and StackMC (blue).
\subsection{Analytic Test Cases}

\subsubsection{Ten-dimensional Rosenbrock function under a uniform $p(x)$, polynomial fitter.}
The D-dimensional Rosenbrock function is given by

\begin{equation}
\label{eq:Rosen}
f(x)=\sum_{i=1}^{D-1}{[(1-x_i)^2+100(x_{i+1}-x_i^2)^2]} \quad \text{,}
\nonumber
\end{equation}
and $x$ is drawn from a uniform distribution over the [-3,3] hypercube.

The fitting algorithm is chosen as a third-order polynomial in each dimension whose form is
\begin{equation}
\nonumber
g(x)= \beta_0+\sum_{i=1}^D \beta_{1,i} x_i+\beta_{2,i} x_i^2+ 
\beta_{3,i} x_i^3 \quad \text{,}
\end{equation}
where the $\beta_i$ are free parameters that are set from the data samples.

\begin{figure}
\centering
\includegraphics[width=0.8\textwidth]{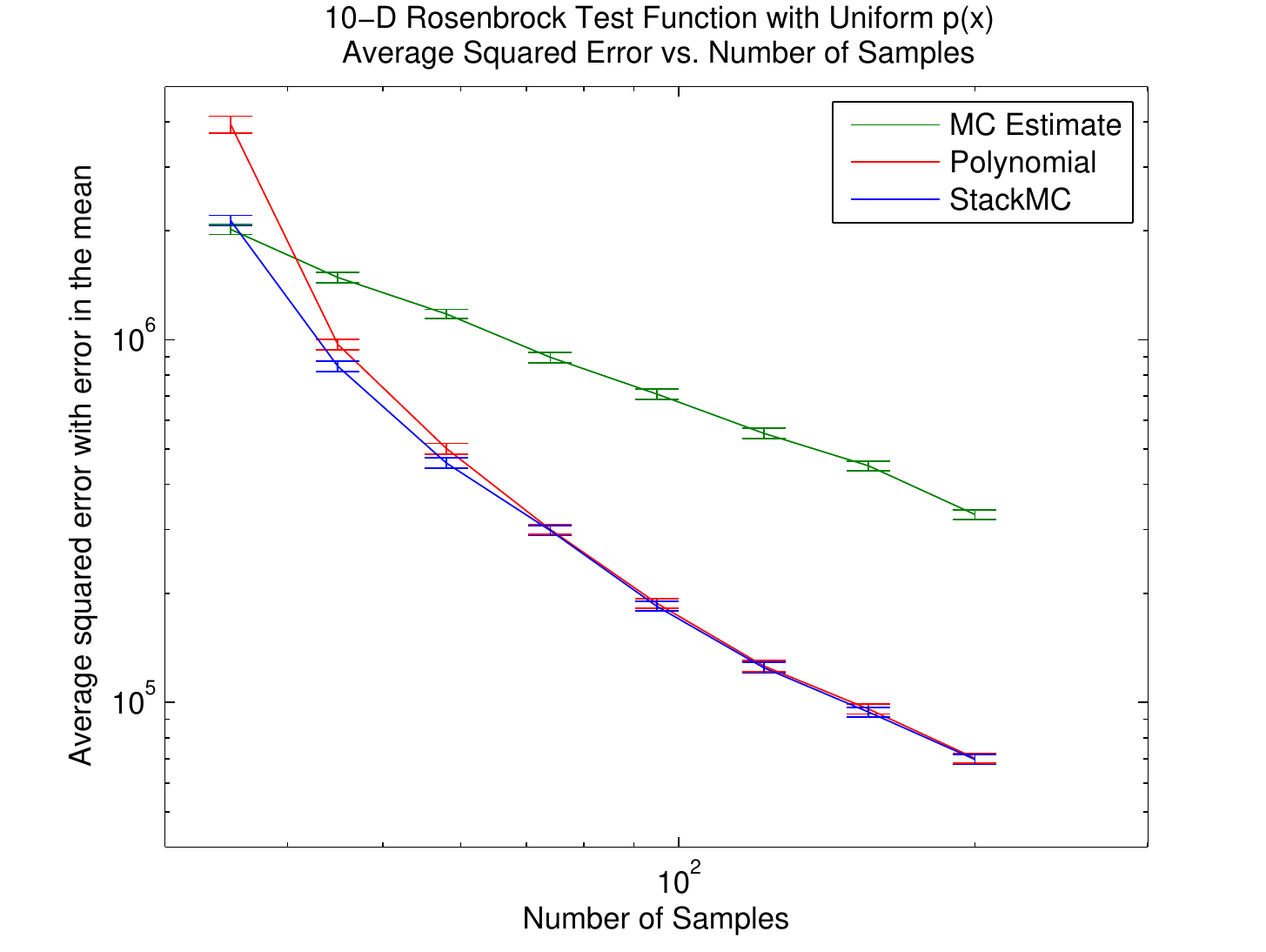}
\caption{Comparison of MC, fitter, and StackMC for the Rosenbrock function with uniform uncertainty.}
\label{fig:tenDRU}
\end{figure}

\begin{figure}
\centering
\includegraphics[width=0.8\textwidth]{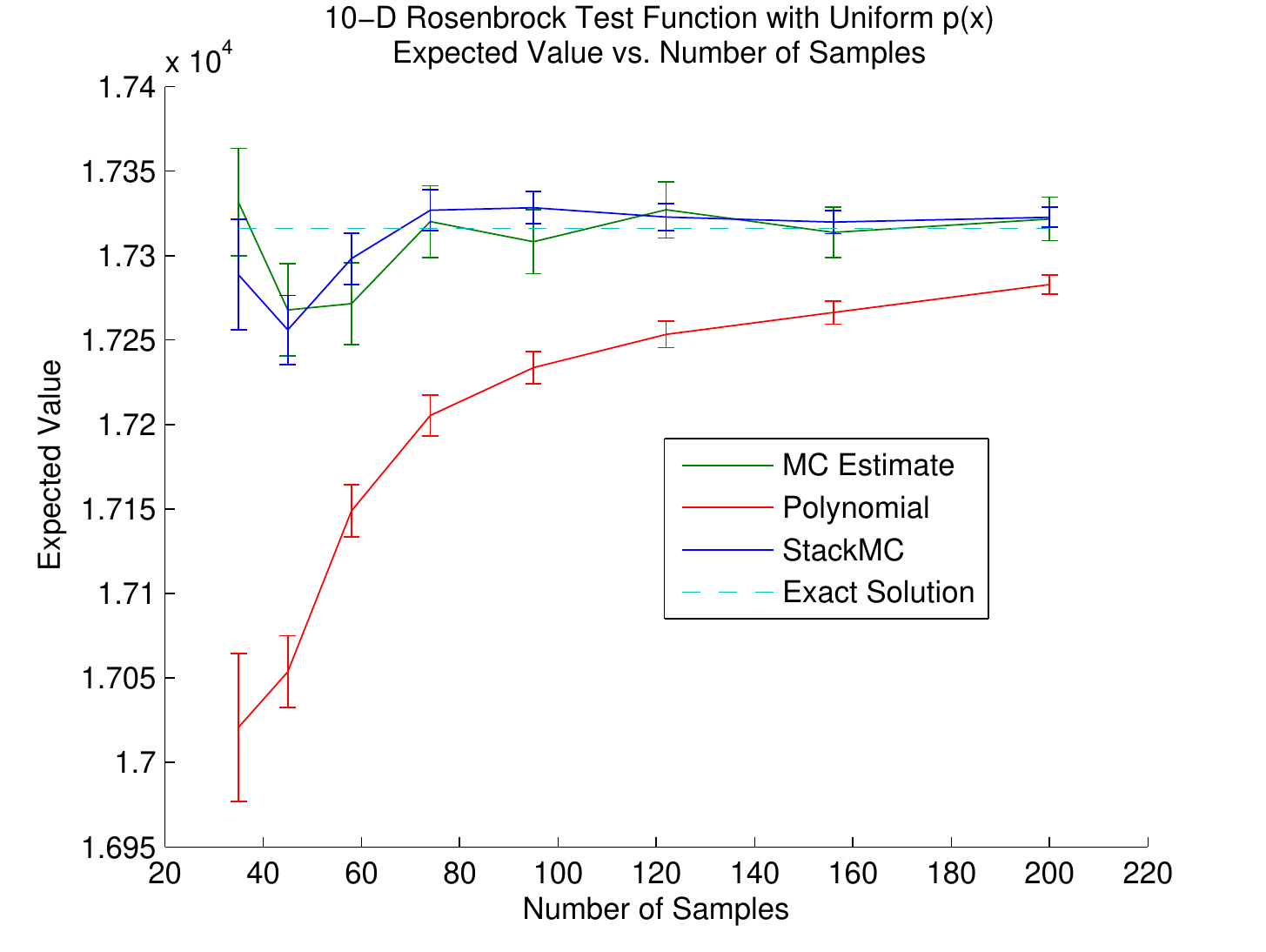}
\caption{Expected output of MC, fitter, and StackMC with error in the mean for the 10-D Rosenbrock function. Note that the fitting function is biased, especially at lower numbers of sample points, while StackMC and MC are not.}
\label{fig:10DRosenUniformEV}
\end{figure}

A comparison of the error of StackMC can be seen in Fig.~\ref{fig:tenDRU} for the ten-dimensional version of the Rosenbrock function. At low numbers of sample points, Monte Carlo is more accurate, on average, than the polynomial fit to all the data points, but at higher numbers of points the polynomial outperforms Monte Carlo. Throughout the entire range of number of samples, StackMC performs at least as well as the best of the two. Additionally, as can be seen in Fig.~\ref{fig:10DRosenUniformEV}, the polynomial fitter is actually a biased fitter for this example problem. StackMC is able to use cross-validation to remove the bias of the fitter while keeping the accuracy of its estimate.

\subsubsection{Ten dimensional Rosenbrock function under a Gaussian $p(x)$.}

This is the same set-up as above, except that $x$ is generated according to a multivariate Gaussian distribution, each dimension independent with $\mu= 0$ and $\sigma=2$.

\begin{figure}
\centering
\includegraphics[width=0.8\textwidth]{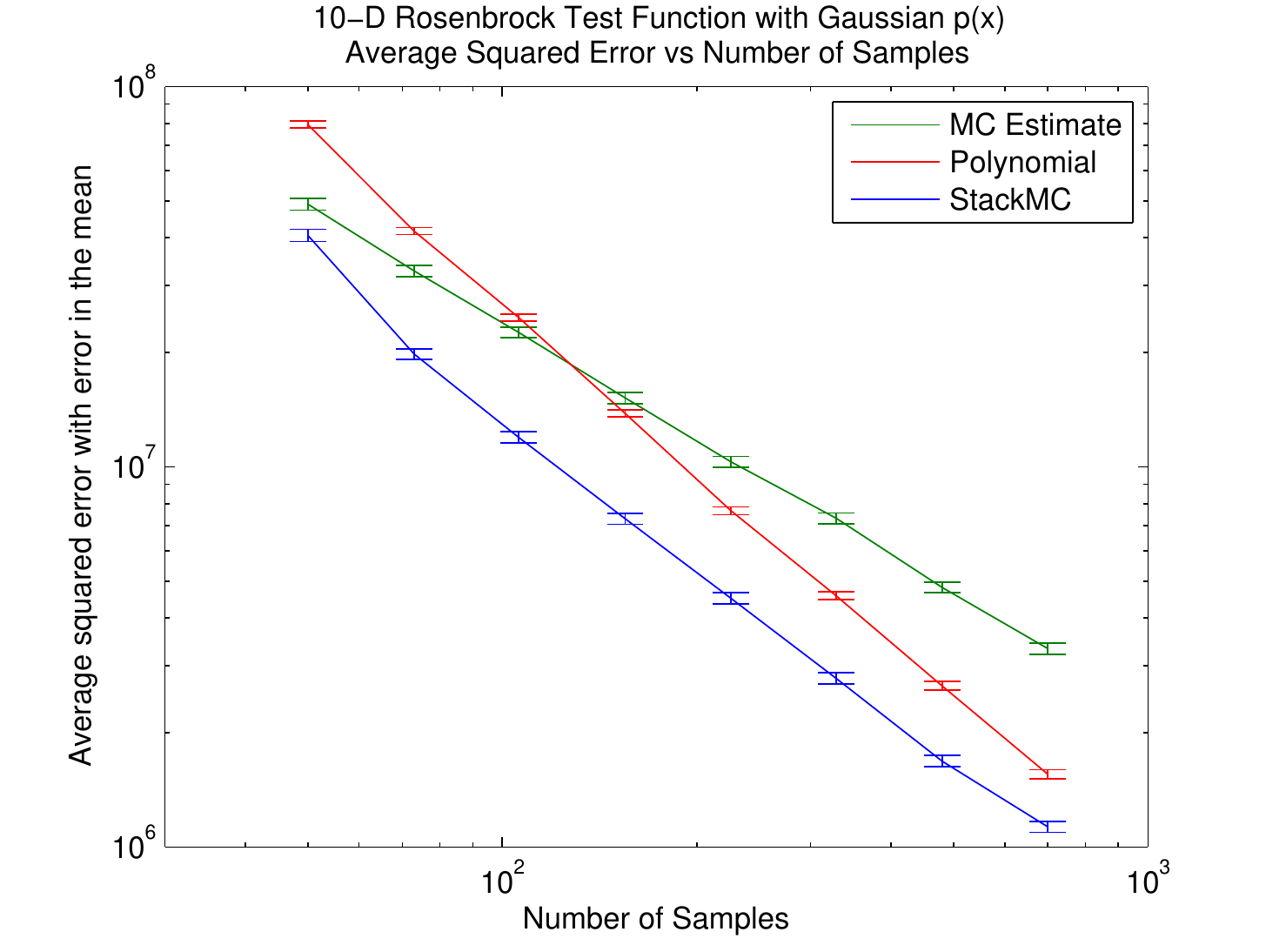}
\caption{Comparison of MC, fitter, and StackMC for the 10-D Rosenbrock function with Gaussian uncertainty.}
\label{fig:10DRosenGauss}
\end{figure}

Much like the case above, in Fig. \ref{fig:10DRosenGauss} it can be seen that at low numbers of sample points Monte Carlo outperforms the polynomial fit, whereas at high sample points the polynomial does better than Monte Carlo alone. However, for all numbers of sample points, StackMC outperforms the other two algorithms.

\subsubsection{20 Dimensional BTButterfly, uniform $p(x)$, Fourier fitter}

In the previous examples, the Rosenbrock function is a 4$^{th}$ order polynomial, and the fitting algorithm is a 3$^{rd}$ order polynomial, so StackMC had reasonable fits to use to improve upon Monte Carlo. In order to challenge StackMC, a function we call the BTButterfly was created so that it would be very difficult to fit accurately.



%

\begin{figure}
\centering
\includegraphics[width=0.8\textwidth]{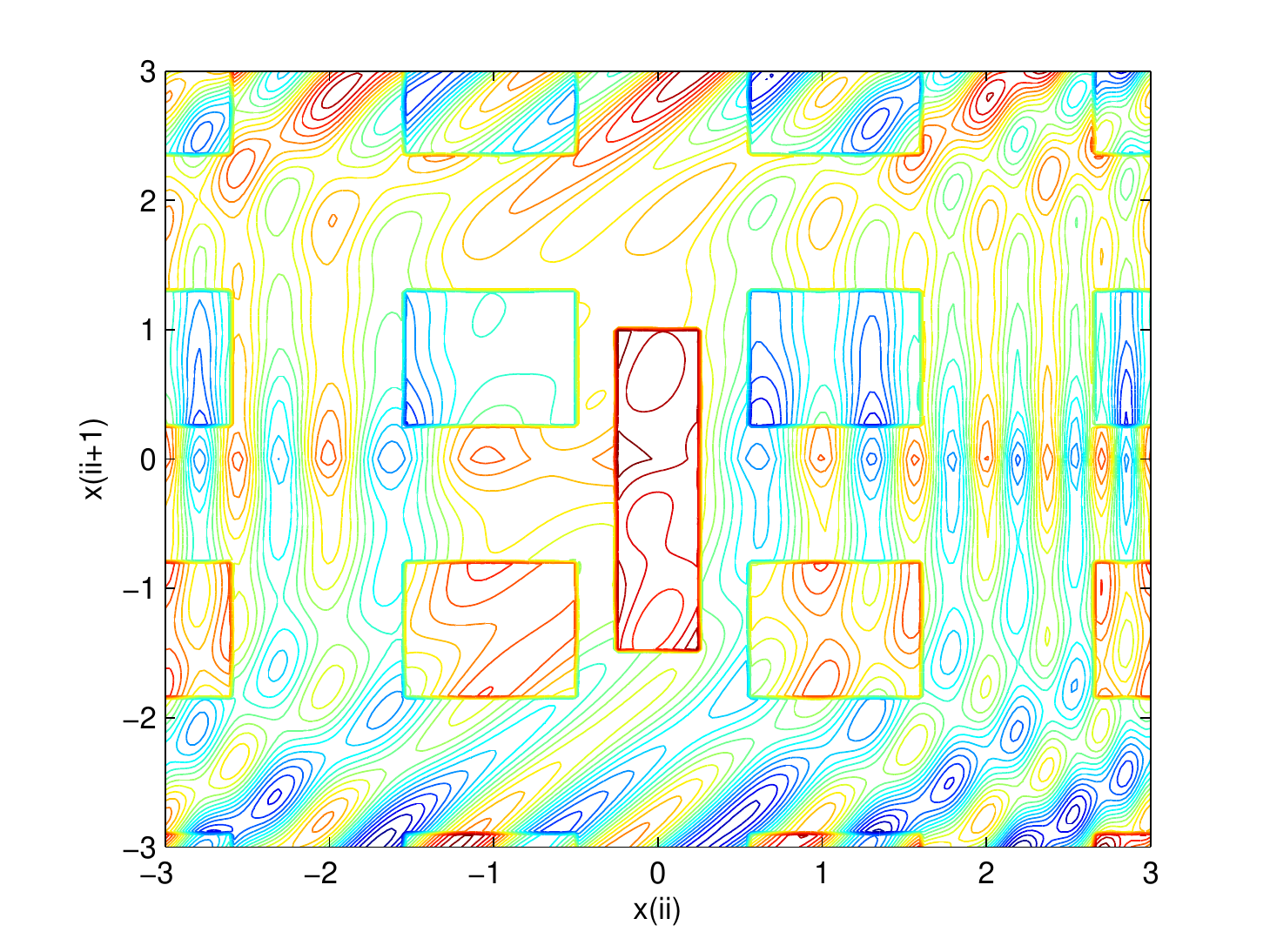}
\caption{Contour plot of successive dimensions of the BTButterfly function.}
\label{fig:BTButterfly}
\end{figure}

Like the Rosenbrock function, its general form is given by
\begin{equation}
\nonumber
\label{eq:BTButterfly}
f(x)=\sum_{i=1}^{D-1}{h(x_i,x_{i+1})}
\end{equation}
with the contour plot of $h$ shown in Fig. \ref{fig:BTButterfly}. $x$ is generated uniformly over the [-3,3] hypercube.

\begin{figure}
\centering
\includegraphics[width=0.8\textwidth]{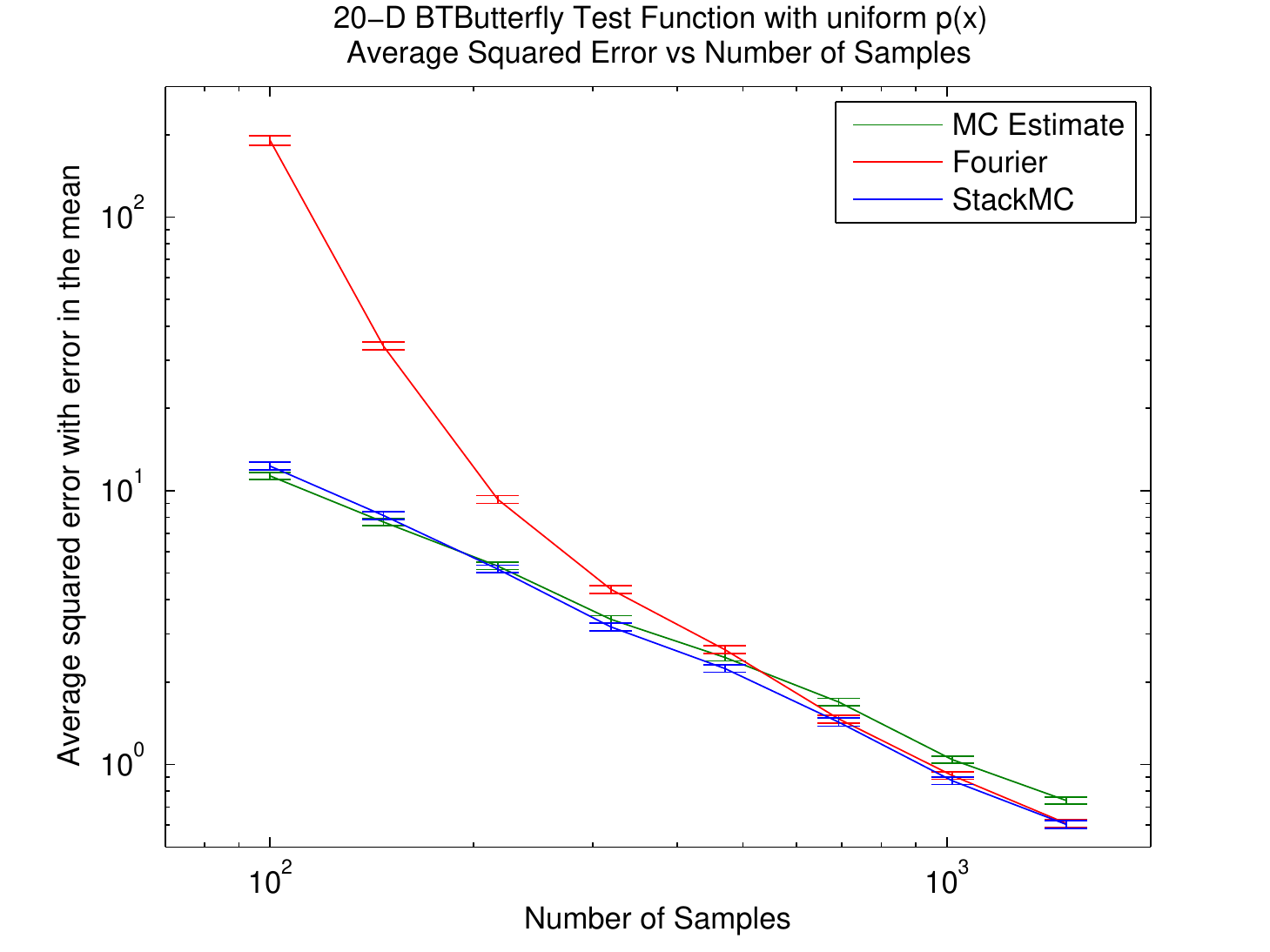}
\caption{Comparison of MC, fitter, and StackMC for the BTButterfly function with uniform uncertainty.}
\label{fig:BTButterflyResults}
\end{figure}

A Fourier expansion for $g(x)$ was chosen whose form is given by

\begin{align}
\nonumber
g(x)= \beta_0+
\sum_{i=1}^D  \beta_{1,i} \cos(x_i)+ \beta_{2,i} \cos(2x_i)+
\beta_{3,i} \cos(3x_i)
+\beta_{4,i} \sin(x)_i+ \beta_{5,i} \sin(2x_i)+
\beta_{3,i} \sin(3x_i) \quad \text{.}
\end{align}
Results for this function are shown in Fig.\ref{fig:BTButterflyResults}, and exhibit the same trends seen previously; StackMC has as low of an error as the lowest of the two methods individually.

\subsection{Aerospace Applications}
\subsubsection{Future Aircraft Uncertainty Quantification (UQ)}
In Ref.\cite{TomSean}, the authors use the Program for Aircraft Synthesis Studies \cite{PASS} (PASS), a conceptual aircraft design tool, to predict the fuel burn of future aircraft given certain assumptions about technology advancement in the 2020 and 2030 time frames. In their predictions for single-aisle aircraft in 2020, the authors model eight probabilistic variables representing different effects of improvements in aircraft technology (propulsion, structures, aerodynamics). Each of the variables is represented by a unique beta distribution. The authors generated 15,000 samples (each representing one optimized aircraft) from which they measured the expected fuel-burn metric and the standard deviation of the expected fuel-burn metric.

To apply StackMC, we choose a third-order polynomial fitter. A polynomial fitter is convenient for this case because the $\E[x^n]$ over a beta distribution is easily found analytically as follows:

\begin{equation}
\E[x^n]=\frac{\prod_{i=1}^n{\alpha +i-1}}{\prod_{i=1}^n{\alpha+\beta +i-1}} \quad \text{,}
\end{equation}
where $\alpha$ and $\beta$ are the two parameters of the beta distribution, $B(\alpha,\beta)$ (not to be confused with the StackMC parameter $\alpha$).

\begin{figure}
\centering
\includegraphics[width=0.8\textwidth]{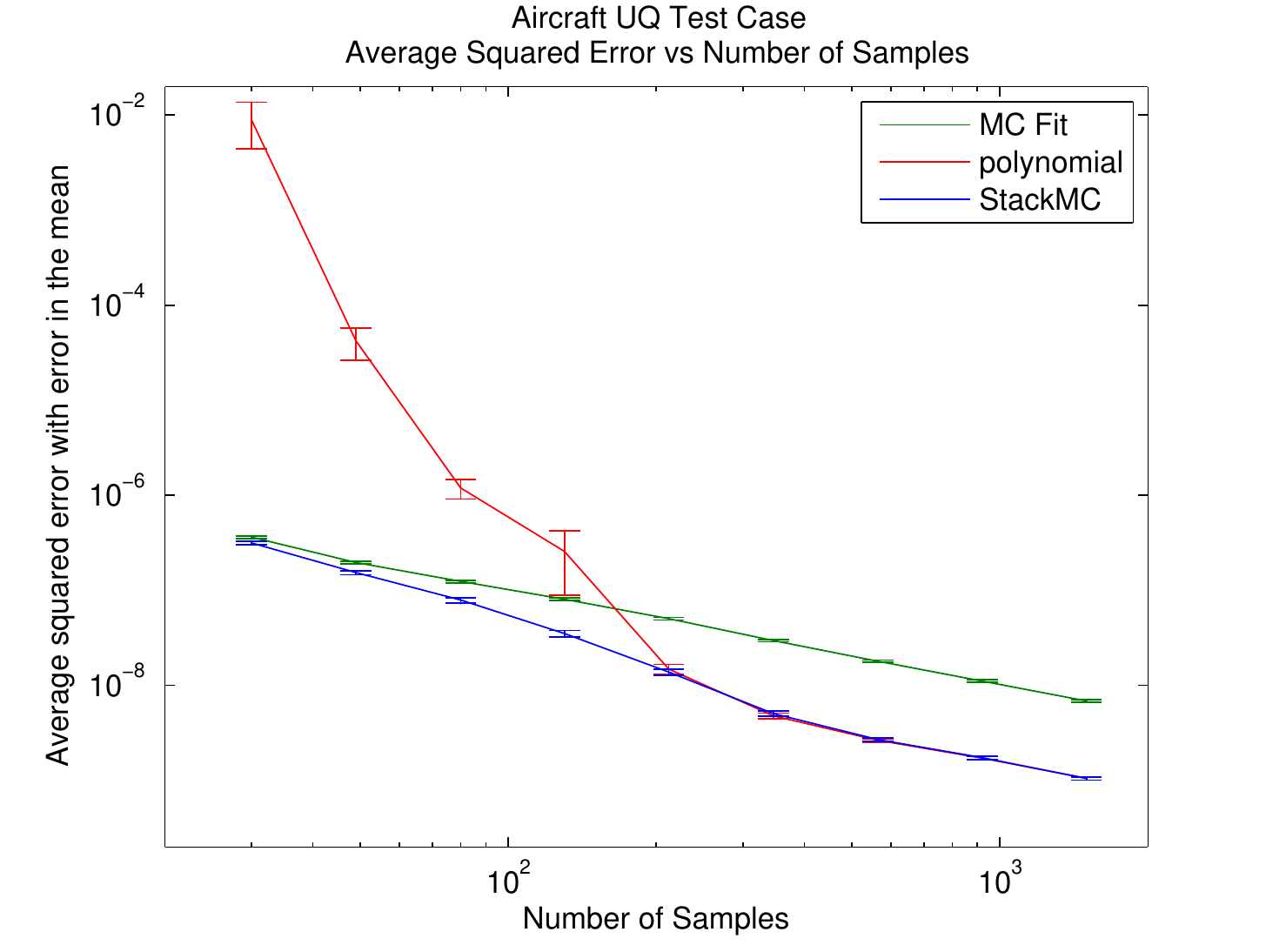}
\caption{Comparison of MC, fitter, and StackMC for the Aircraft UQ test case finding $\E[x]$.}
\label{fig:AircraftUQEV}
\end{figure}

\begin{figure}
\centering
\includegraphics[width=0.8\textwidth]{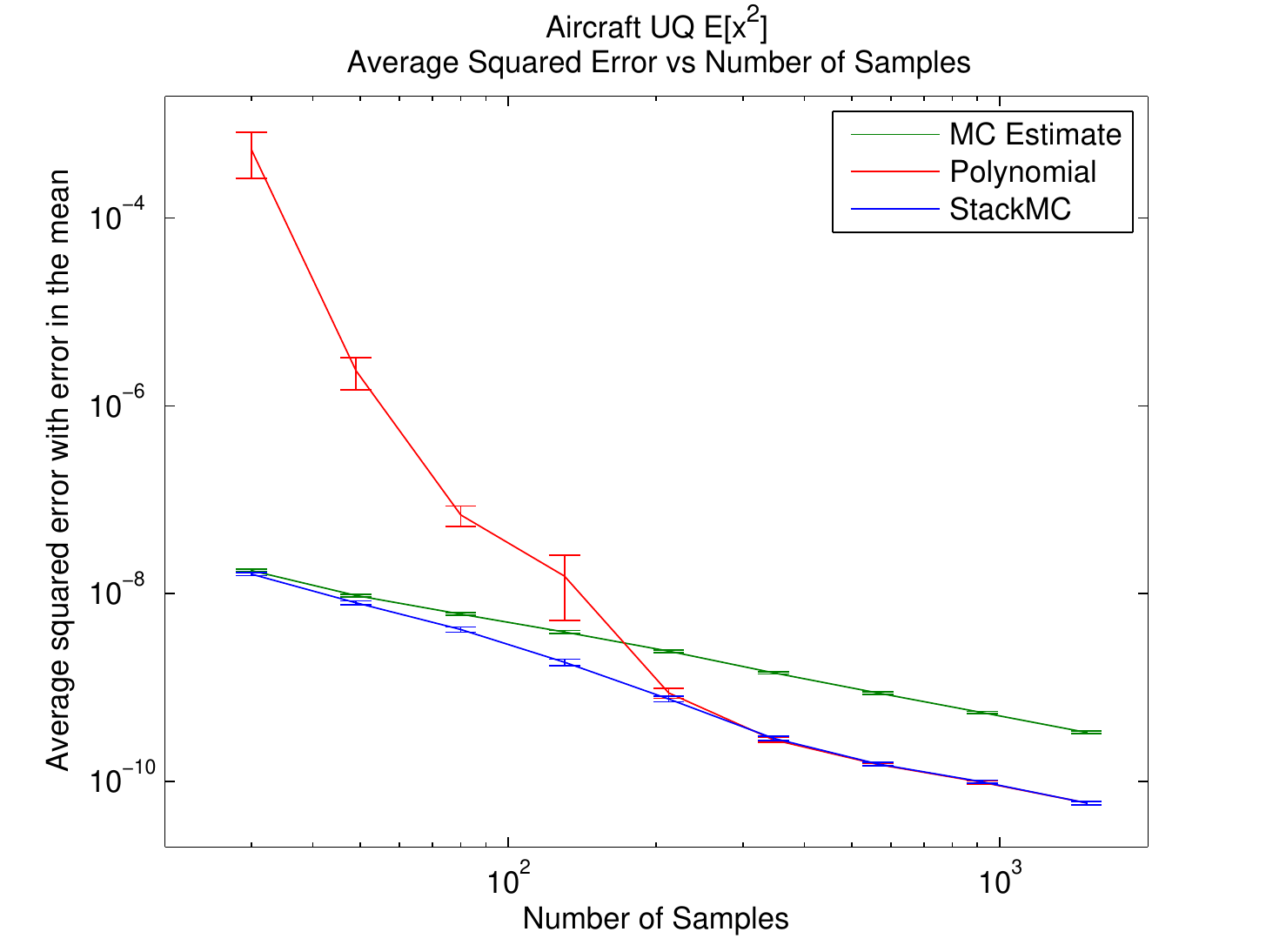}
\caption{Comparison of MC, fitter, and StackMC for the Aircraft UQ test case finding $\E[x^2]$.}
\label{fig:AircraftUQEX2}
\end{figure}

A set of 100,000 samples were generated and the mean and standard deviation of their function values were taken as the ``true'' expected value and standard deviation. The standard deviation is defined as $\sqrt{\E[x^2]-\E[x]^2}$, and therefore, to find the standard deviation we can run StackMC twice, once to find $\E[x]$ and a second time to find $\E[x^2]$. The results of applying StackMC to find the expected value and standard deviation are shown in Fig.~\ref{fig:AircraftUQEV} and Fig.~\ref{fig:AircraftUQEX2}, respectively.

The exact same trend is seen here as in all of the analytic problems. At low numbers of samples, where the fit to all the data points performs badly, StackMC does as well as Monte Carlo. At high numbers of samples, where the fit greatly outperforms Monte Carlo, StackMC does as well as the fitter. In between these two extremes, StackMC outperforms both algorithms.

At 1500 samples, StackMC reduces the error by roughly an order of magnitude compared with the Monte Carlo result used by the authors. Each sample takes about 6 seconds to generate, whereas StackMC takes less than a half a second to form the 10 fits, calculate $\alpha$, and calculate \eqref{eq:smc1} . For a computational cost of less than one additional sample, StackMC has reduced the number of samples needed by 90\% for the same expected error.

\subsubsection{Sonic Boom Uncertainty Quantification}
\begin{figure}
\centering
\includegraphics[width=0.8\textwidth]{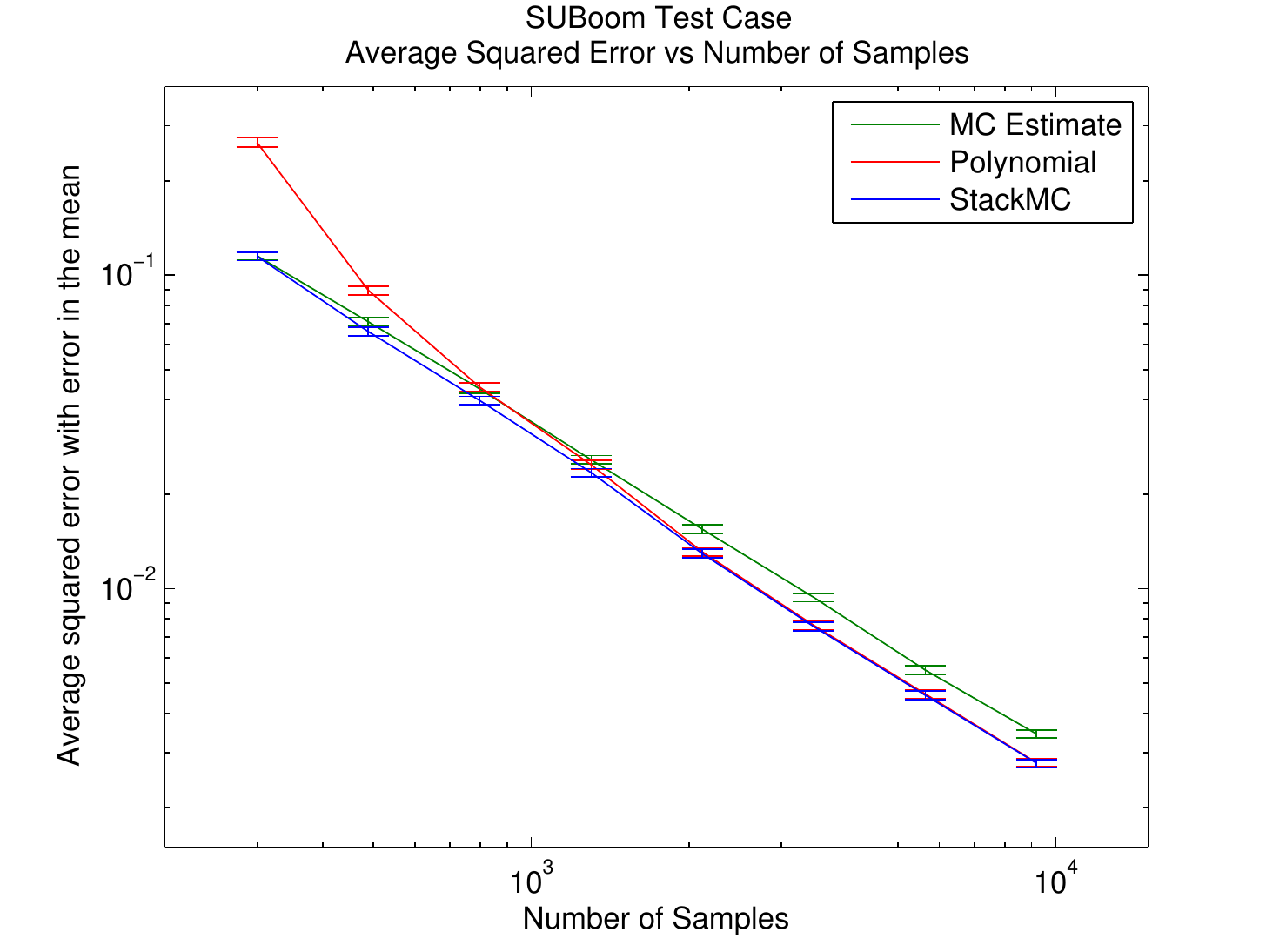}
\caption{Comparison of MC, fitter, and StackMC for the SUBoom UQ test case.}
\label{fig:SUBoomUQ}
\end{figure}

The uncertainty quantification of sonic boom noise is used as the second application case. Unlike the aircraft design test case, the response surface for the sonic boom noise signature is not smooth; the output can vary significantly with slight adjustments to the input parameters. Colonno and Alonso\cite{SUBoom} recently created a new sonic boom propagation tool, SUBoom, and used it to analyze the robustness of several aircraft pressure signatures optimized for minimal boom noise. Specifically, in their high-fidelity near-field case, they have 62 uncertain variables: 4 representing aircraft parameters such as cruise Mach number and roll angle, and 58 representing uncertainties in the near-field pressure signal. Like in the Aircraft UQ case, 100,000 samples were generated and used as the true mean. A third-order polynomial was used as $g(x)$. The results can be seen in Fig. \ref{fig:SUBoomUQ}.

Even with the discontinuities in the function space, the same performance for StackMC is seen. StackMC does as well as the best of Monte Carlo and the fitting algorithm. The reduction in error is not as great here as in the Aircraft UQ case because a polynomial is not a good fit to $f(x)$, though using domain knowledge to improve the accuracy of the fitting algorithm would further increase the performance of StackMC. Despite the relatively poor performance of the fitting algorithm, it is still better to use StackMC than to not regardless of the number of sample points used. In some senses, this test case demonstrates one of the strengths of StackMC: at virtually no additional computation cost the user is guaranteed the highest accuracy without having to worry about the appropriateness of the methodology for a given number of samples.

\section{Conclusion}

In this paper, we have introduced a new technique, Stacked Monte Carlo, which reduces error of Monte Carlo sampling by blending several different fitters of $f(x)$. StackMC is unbiased, (thus retaining a major advantage of Monte Carlo), and is able to use cross validation to effectively incorporate the use of a fitting function. 

StackMC is an extremely generic post-processing technique. It is applied after the samples are generated and makes no assumptions about the generation method, and therefore StackMC can be used not only with simple MC, but also with other sample generation techniques such as Importance Sampling, Quasi-Monte Carlo, and Markov-Chain Monte Carlo. Furthermore, StackMC makes no assumptions about $f(x)$; it not only applies to smooth functions, but also to discontinuous functions, and even functions with discrete variables. In a future paper, \cite{futurepaper} we will present results of the application of StackMC to different input regimes and different sample generation methods. We will examine higher dimensional spaces, explore the application of StackMC to multi-fidelity methods, and extended StackMC to incorporate multiple fitting functions.

The computation time of StackMC is dominated by forming the fits $g_i(x)$. As a result, StackMC is only affected by the curse of dimensionality insofar as the fitting algorithm is. In both of the aerospace applications, the additional cost of the entire StackMC algorithm was virtually non-existant; there are significant improvements in accuracy for less computation time than the cost of one additional function evaluation. The only assumptions made about $g(x)$ are that it be able to predict the value at new sample locations and we are able to evaluate $\ghat$ analytically (and even this second assumption is relaxed). As shown in the BTButterfly and sonic boom example cases, the fit does not need to be particularly good to see improvement, and so the choice for $g(x)$ can be modified as computational effort and domain knowledge allows. StackMC does not eliminate the need for finding better fitters; a better fitter will always lead to an improved result.  With the exception of the error in the mean test (whose value we did not change for any of the example cases), StackMC requires no user-set parameters that need to be heuristically tuned to see good results. 

The version of StackMC implemented in this paper is relatively naive. Instead of taking the mean of the results from the different fits, taking a weighted combination of the fits could improve the estimate. StackMC currently only partitions the data into folds once, but we could imagine re-partitioning the same samples several times to have more fitters. We use Monte Carlo to estimate the integral term in \eqref{eq:fitterint}, but using a fitter, or even using StackMC recursively, could improve the estimates of $\hat{f}_{smc}$. Furthermore, $\alpha$ was set as a constant, but in general $\alpha$ could vary between the folds or could even be a function of $x$ so the confidence in the fit varies spatially.

Despite this simplistic implementation, StackMC performs at least as well as the better of MC and the fitting function, and for a range of sample points outperforms both. Normally, there is a transition number of samples at which using a fit to all the data samples has a smaller average error than MC, and it is hard to know \emph{a priori} if it is better to use the fit. StackMC eliminates the need to decide whether or not to use a fit; it will never be harmful to do so. While it is not true that for any set of data samples $\fhatsmc$ is closer to $\fhat$ than $\fhatmc$, on average StackMC reduces the expected error. StackMC is a very generic method for the post-processing of data samples; it can be used by anyone trying to estimate an integral or expected value of a function where $p(x)$ is known. 


\bibliographystyle{aiaa}
\bibliography{StackMC_AIAA_Paper}

\end{document}